\documentclass[sigconf]{acmart}

\usepackage{graphicx}
\usepackage{algorithm}
\usepackage{algorithmic}
\usepackage{subfigure}
\usepackage{array}
\usepackage{setspace}
\usepackage{float}
\usepackage{url}
\usepackage{balance}
\usepackage{amsfonts}
\usepackage{amsmath}
\usepackage{rotating}
\usepackage{multirow}
\usepackage{color}
\usepackage{bm}
\usepackage{enumitem}
\usepackage{pifont}
\usepackage[T1]{fontenc}
\usepackage{colortbl}
\definecolor{bleudefrance}{rgb}{0.19, 0.55, 0.91}

\newcommand{\model}{MATCH}

\newcommand{\bfH}{{\bm H}}
\newcommand{\bfe}{{\bm e}}
\newcommand{\bfq}{{\bm q}}
\newcommand{\bfa}{{\bm a}}
\newcommand{\bfK}{{\bm K}}
\newcommand{\bfV}{{\bm V}}
\newcommand{\bfW}{{\bm W}}
\newcommand{\bfz}{{\bm z}}
\newcommand{\bfh}{{\bm h}}
\newcommand{\bfpi}{{\bm \uppi}}
\newcommand{\bfb}{{\bm b}}
\newcommand{\bfw}{{\bm w}}
\newcommand{\bfy}{{\bm y}}
\newcommand{\bfc}{{\bm c}}

\newcommand{\cmark}{\ding{51}}
\newcommand{\xmark}{\ding{55}}

\definecolor{darkg}{rgb}{0.0, 0.5, 0.0}

\newcommand{\hide}[1]{} 

\newtheorem{problem}{Problem}

\makeatletter
\def\@fnsymbol#1{\ensuremath{\ifcase#1\or \dagger\or \ddagger\or
   \mathsection\or \mathparagraph\or \|\or **\or \dagger\dagger
   \or \ddagger\ddagger \else\@ctrerr\fi}}
\makeatother

\copyrightyear{2021}
\acmYear{2021}
\setcopyright{iw3c2w3}
\acmConference[WWW '21]{Proceedings of the Web Conference 2021}{April 19--23, 2021}{Ljubljana, Slovenia}
\acmBooktitle{Proceedings of the Web Conference 2021 (WWW '21), April 19--23, 2021, Ljubljana, Slovenia}
\acmPrice{}
\acmDOI{10.1145/3442381.3449979}
\acmISBN{978-1-4503-8312-7/21/04}

\begin{document}

\title{\model: Metadata-Aware Text Classification in \\ A Large Hierarchy}

\author{Yu Zhang}
\authornote{Work performed while interning at Microsoft Research.}
\affiliation{
\institution{Univ. of Illinois at Urbana-Champaign}
\institution{yuz9@illinois.edu}
}

\author{Zhihong Shen}
\affiliation{
\institution{Microsoft Research, Redmond}
\institution{zhihosh@microsoft.com}
}

\author{Yuxiao Dong}
\authornote{Now at Facebook AI, Seattle and work done while working at Microsoft Research.}
\affiliation{
\institution{Microsoft Research, Redmond}
\institution{ericdongyx@gmail.com}
}

\author{Kuansan Wang}
\affiliation{
\institution{Microsoft Research, Redmond}
\institution{kuansanw@microsoft.com}
}

\author{Jiawei Han}
\affiliation{
\institution{Univ. of Illinois at Urbana-Champaign}
\institution{hanj@illinois.edu}
}

\begin{abstract}

Multi-label text classification refers to the problem of assigning each given document its most relevant labels from a label set. 
Commonly, the metadata of the given documents and the hierarchy of the labels are available in real-world applications. 
However, most existing studies focus on only modeling the text information, with a few attempts to utilize either metadata or hierarchy signals, but not both of them. 
In this paper, we bridge the gap by formalizing the problem of metadata-aware text classification in a large label hierarchy (e.g., with tens of thousands of labels). 
To address this problem, we present the \textsf{\model}\footnote{The code and datasets are available at {\color{bleudefrance} \url{https://github.com/yuzhimanhua/MATCH}}.} solution---an end-to-end framework that leverages both metadata and hierarchy information. 
To incorporate metadata, we pre-train the embeddings of text and metadata in the same space and also leverage the fully-connected attentions to capture the interrelations between them. 
To leverage the label hierarchy, we propose different ways to regularize the parameters and output probability of each child label by its parents.
Extensive experiments on two massive text datasets with large-scale label hierarchies demonstrate the effectiveness of \textsf{\model} over the state-of-the-art deep learning baselines.

\end{abstract}



\maketitle

\section{Introduction}
\begin{figure}
\centering
\subfigure[Input: a PubMed paper with metadata]{
\includegraphics[scale=0.29]{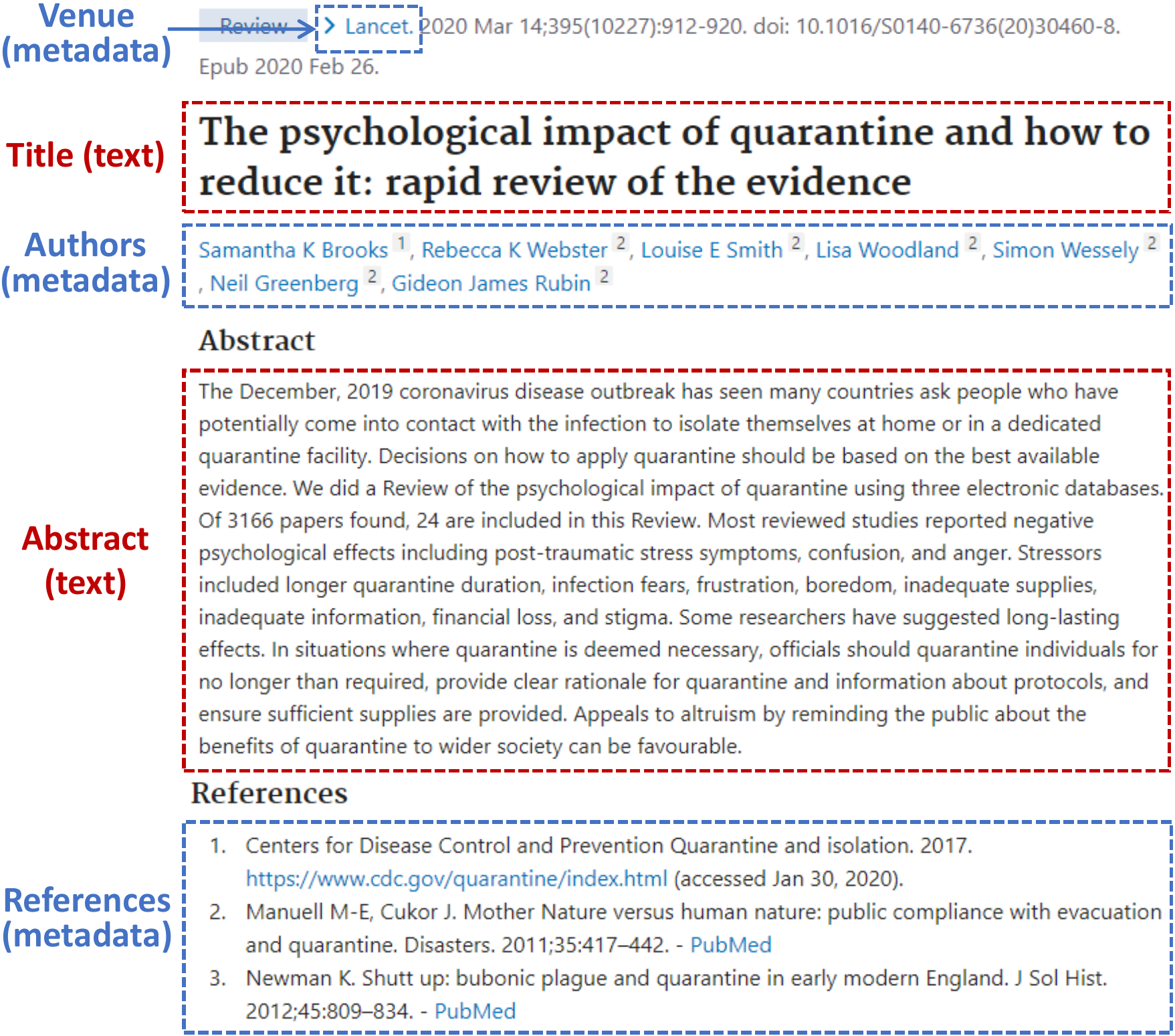}}
\hspace{-0.5em}
\subfigure[Input: MeSH hierarchy]{
\includegraphics[scale=0.29]{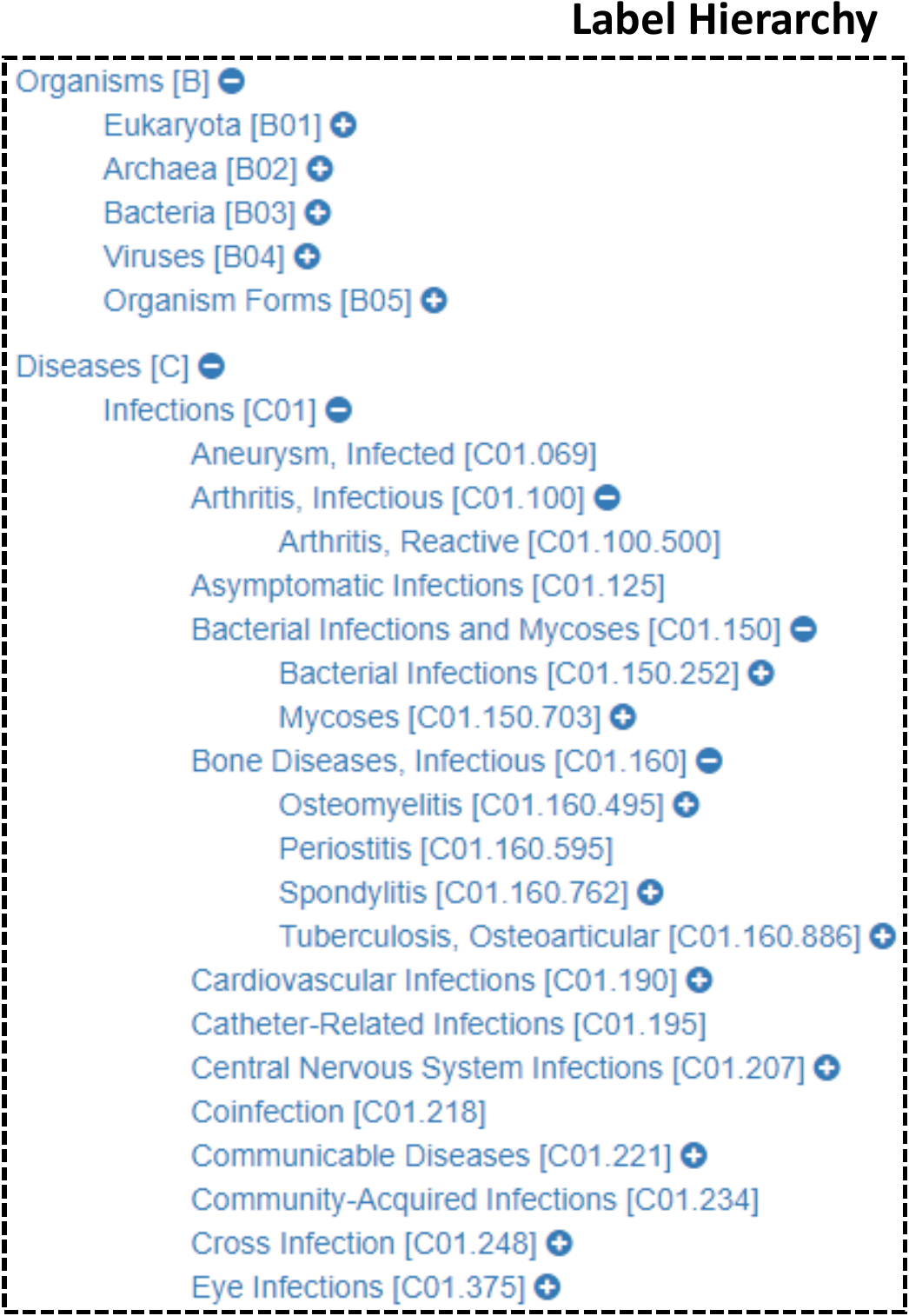}}
\vspace{-0.5em}
\subfigure[Relevant labels (MeSH terms) of the document]{
\includegraphics[scale=0.285]{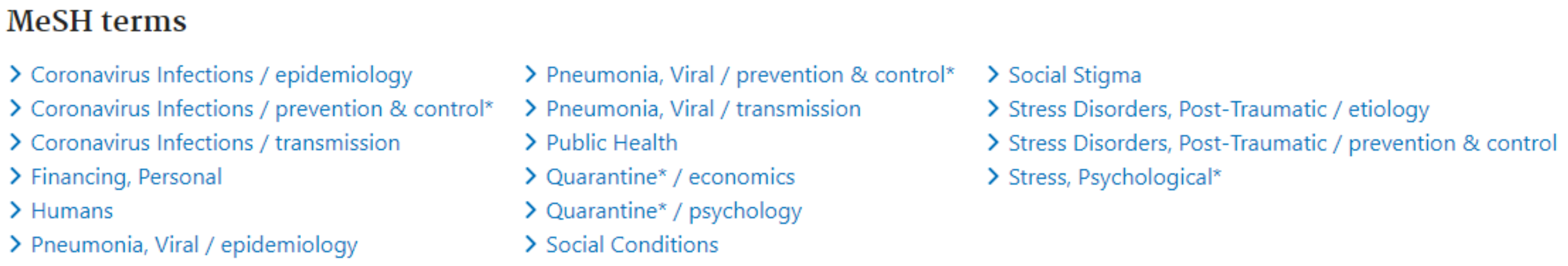}}
\caption{An example of metadata-aware hierarchical text classification on PubMed. We utilize both (a) the metadata of documents and (b) a large-scale label hierarchy to predict (c) relevant labels of each document.} 
\vspace{-0.5em}
\label{fig:example}
\end{figure}

Text classification is a fundamental text mining task \cite{aggarwal2012mining}.   
In the age of information overload, it becomes particularly important as the exponential growth of accessible documents. 
Take the science enterprise as an example, the volume of publications has doubled every 12 years \cite{dong2017century}, reaching in total 240,000,000 by 2019 \cite{wang2020microsoft}, and 
by February 2021, 213,236 papers on COVID-19\footnote{\url{https://academic.microsoft.com/topic/3008058167/}, accessed on Feb. 12, 2021.} had already been generated.  
This explosion in publications makes the mission of tracking the related literature impossible, requiring accurate classification of them into different levels of topics more than ever. 

The current attempt to address this problem is mainly focused on leveraging the power of deep neural networks, such as the CNN based XML-CNN model \cite{liu2017deep} and the RNN based AttentionXML model \cite{you2019attentionxml}. 
More recently, X-Transformer \cite{chang2020taming}---a pre-trained language model based technique---is presented to perform large-scale text classification. 
However, the majority of these studies only model the text information of documents and are less concerned with two widely-available signals in real-world applications: \emph{document metadata} and \emph{a large-scale label hierarchy}.

To illustrate the scenario, Figure \ref{fig:example} takes a scientific paper on PubMed as an example.  
We can see that, in addition to its text information (title and abstract), the paper is also associated with various types of metadata, such as its publication venue, authors, and references, which could be strong indicators of its research topics. 
For instance, its venue ``\textit{Lancet}'' would strongly suggest the paper is most likely related to medicine; the first three publications it cites would further indicate the paper's relevance to epidemiology. 
Broadly, metadata is also commonly available for other digitized documents, such as online posts, product reviews, and code repositories. 
However, this common information is largely unexplored in existing studies \cite{liu2017deep,you2019attentionxml,chang2020taming}. 

Furthermore, research topics on PubMed are organized in a hierarchical way, such as the parent topic of ``\texttt{Infections}'' is ``\texttt{Diseases}'' and one of its child topics is ``\texttt{Eye Infections}'', providing signals that are not offered in text alone. 
For example, the hierarchy suggests the high prediction confidence in ``\texttt{Eye Infections}'' for one paper is also a strong indicator of being ``\texttt{Infections}'' related. 
Consequently, it can also benefit topics with sparse training data. 
Though most label systems for text data are naturally organized into hierarchies, such as web directories \cite{partalas2015lshtc} and product catalogs \cite{mao2020octet}, this signal has often been left out \cite{liu2017deep,you2019attentionxml,chang2020taming} or used in a small label space \cite{peng2018large,zhou2020hierarchy,zhang2021hierarchical}.

\vspace{1mm}

\noindent \textbf{Contributions.}
To bridge the gap, we formalize the problem of metadata-aware text classification in a large-scale label hierarchy. 
Specifically, given a collection of documents, the task is to train a multi-label classifier that incorporates not only their text information but also both the metadata and taxonomy signals for inferring their labels. 
To address this problem, we present the \textsf{\model} framework that fully utilizes both signals. 
To exploit the metadata of input documents, we propose to generate the pre-trained embeddings of text (i.e., words) and metadata in the same latent space. 
We further leverage the fully connected attention mechanism in Transformer to capture all pairwise relationships between words and different types of metadata, which produces an expressive representation for each document with its metadata encoded. 
Empirical evidence suggests that the modeling of metadata not only helps improve the classification results but also accelerates the convergence of classifier training. 

To incorporate the label hierarchy, we design strategies to regularize the parameters and output probability of each child label by its parents. 
In the parameter space, we encourage the child and parent labels to have similar parameters in the prediction layer, that is, determining whether a document would be tagged with a child label (e.g., ``\texttt{Eye Infections}'') should share similarities with whether to assign it with its parent (e.g., ``\texttt{Infections}'').  
In the output space, we introduce a regularization inspired by the distributional inclusion hypothesis \cite{geffet2005distributional}. 
Intuitively, it requires the probability that a document belongs to a parent label to be no less than the ones that it is associated with its children. 
Such a regularization strategy characterizes the asymmetric hypernym-hyponym relationship, which is beyond the symmetric similarity in the parameter space.

Empirically, we demonstrate the effectiveness of \textsf{\model} on two massive text datasets extracted from the Microsoft Academic Graph \cite{sinha2015overview,wang2019a} and PubMed \cite{lu2011pubmed}. 
Both datasets contain large-scale topic hierarchies with more than 15K labels.
The results suggest that \textsf{\model} can consistently outperform the state-of-the-art multi-label text classification approaches as well as Transformer-based models. 
Moreover, we validate the design choices of incorporating metadata and the label hierarchy for text classification. 
Finally, we present several case studies to illustrate how \textsf{\model} specifically benefits from these two sets of signals. 

To summarize, this work makes the following contributions: 
\begin{itemize}[leftmargin=*]
\item We formalize the problem of text classification with the metadata of documents and a large-scale hierarchy of labels, which are usually not simultaneously modeled in existing studies.

\item We design an end-to-end \textsf{\model} framework that incorporates both document metadata and a large label hierarchy for the text classification task. 

\item We conduct extensive experiments on massive online text datasets to demonstrate the effectiveness of the proposed \textsf{\model} framework and its design choices.
\end{itemize}

The rest of the paper is organized as follows. We define several concepts and formulate the problem in Section 2. Then, we present the \textsf{\model} framework in Section 3. We conduct experiments in Section 4 and review related work in Section 5. Finally, Section 6 concludes this study.

\begin{figure*}
\centering
\includegraphics[width=\linewidth]{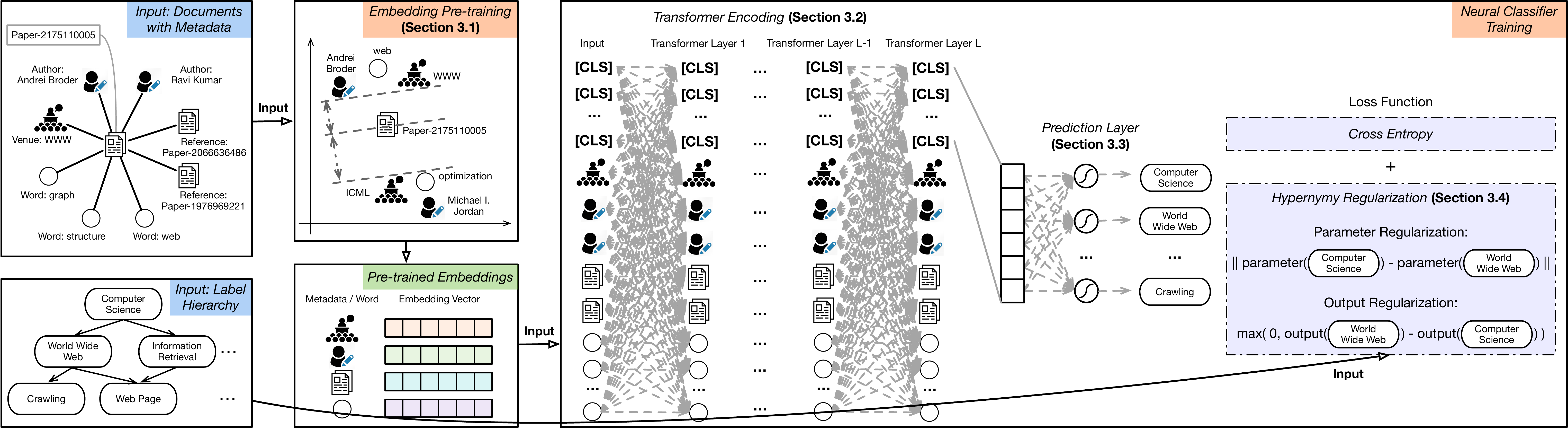}
\vspace{-1em}
\caption{Overview of the \textsf{\model} framework.} 
\vspace{-1em}
\label{fig:framework}
\end{figure*}

\section{Problem Definition}
We study the problem of multi-label text classification. 
Traditionally, this problem is formalized as using only the text information of documents as the input for inferring their labels \cite{liu2017deep,you2019attentionxml,chang2020taming}. 
Here text refers to all free-text fields of a document (e.g., the title and abstract of a scientific publication).

However, the metadata of documents and the hierarchy of labels are usually also available in real-world applications. 
Take the academic publication in Figure \ref{fig:example} as an example, the metadata of one document includes its authors (e.g., ``\textit{Samantha K Brooks}''), published venue (e.g., ``\textit{Lancet}''), and referenced papers. 
The label hierarchy is organized based on the fine-grained levels of research topics, such as ``\texttt{Diseases}'', ``\texttt{Infections}'', and ``\texttt{Eye Infections}''. 

Formally, we can represent the text information of a document $d$ as a single word sequence $\mathcal{W}_d=w_1w_2\cdots w_N$ concatenated from all its text fields, and all its metadata as a 
set $\mathcal{M}_d = \{m_1, m_2, \cdots, m_M\}$.
The label hierarchy can be represented as a tree or a directed acyclic graph (DAG) that specifies the hypernym-hyponym relationships between labels. 
In both cases, the label hierarchy can be characterized by a mapping $\Phi: \mathcal{L} \rightarrow 2^\mathcal{L}$, where $\Phi(l)$ is the set of parent labels of $l\in \mathcal{L}$. If $l$ does not have any parent in $\mathcal{L}$, i.e., $l$ is the root of a tree, we set $\Phi(l) = \emptyset$. 
We formalize the problem of the metadata-aware text classification with a label hierarchy as follows:

\begin{problem}
\textit{
Given a training corpus $\mathcal{D} = \{d_1,...,d_{|\mathcal{D}|}\}$, the label space $\mathcal{L}$ and its hierarchy $\Phi$, where each document $d$ is associated with its text information $\mathcal{W}_d$, metadata $\mathcal{M}_d$, and labels $\mathcal{L}_d \subseteq \mathcal{L}$, the objective is to learn a multi-label classifier $f_{\rm class}$ that maps a document to a subset of $\mathcal{L}$.
}
\end{problem}

Different from the conventional multi-label text classification setting \cite{ji2008extracting,liu2017deep,you2019attentionxml}, the task here is both hierarchy- and metadata-aware. There are some previous approaches which have leveraged metadata into text classification problems, ranging from review sentiment analysis \cite{tang2015learning} and tweet localization \cite{zhang2017rate} to generic classification tasks \cite{kim2019categorical,zhang2020minimally}. However, these studies are all designed for flat text classification. 

Along another line of work, some studies try to utilize the label hierarchy via recursive regularization \cite{gopal2013recursive,gopal2015hierarchical,peng2018large} or hierarchy-aware deep neural models \cite{wehrmann2018hierarchical,mao2019hierarchical,zhou2020hierarchy}. However, these approaches are unaware of the metadata signals accompanying each document.
The main challenge of our task is then how to simultaneously incorporate the metadata of documents and a hierarchy of labels into a unified learning framework.

\section{The \textsf{\model} Framework}

In this section, we present the \textsf{\model} framework for the metadata- and hierarchy-aware multi-label text classification problem.
The overall \textsf{\model} framework is illustrated in Figure \ref{fig:framework}, where we use the paper ``Graph structure in the Web''\footnote{https://academic.microsoft.com/paper/2175110005/} as a running example.
To incorporate the metadata of documents, \textsf{\model} jointly pre-trains the embeddings of metadata, text, and labels into the same latent space, which are further fed into a Transformer model for generating the document representation for prediction. 
To leverage the hierarchy of labels, \textsf{\model} regularizes the parameters and output probabilities of each child label by its parents.

\textsf{\model} can be decomposed into four modules: 
(1) metadata-aware embedding pre-training, (2) Transformer encoding, (3) prediction, and (4) hypernymy regularization.

\subsection{Metadata-Aware Embedding Pre-Training}
Recently, using pre-trained word embeddings \cite{mikolov2013distributed} as the initial input has become a \textit{de facto} standard for training a neural text classifier \cite{kim2014convolutional,yang2016hierarchical,guo2019star}. 
However, in our task, it is also required to capture the relationships between text and its metadata, and preferably to have them embedded in the same latent space. 
To achieve this, we propose a metadata-aware embedding pre-training module to jointly learn their representations by considering several types of proximities between them. 

\vspace{1mm}\noindent \textbf{Document \& Metadata.} 
To preserve the proximity between a document $d$ and its metadata instances $m \in \mathcal{M}_d$ in the joint embedding space, following previous studies on word embedding \cite{mikolov2013distributed} and network embedding \cite{tang2015line}, we define the following conditional probability:
\begin{equation}
    p(m|d) = \frac{\exp(\bfe_m^T\bfe_d)}{\sum_{m' \in \mathcal{V}_m}\exp(\bfe_{m'}^T\bfe_d)},
\label{eqn:pv}
\end{equation}
where $\mathcal{V}_m$ is the set of metadata instances sharing the same type with $m$ (e.g., if $m$ denotes a venue, then $\mathcal{V}_m$ is the set of all venues appearing in the training set); $\bfe_m$ and $\bfe_d$ are metadata and document embedding vectors, respectively.

Given a positive document-metadata pair $(d, m_+)$, our goal is to maximize the log-likelihood $\log p(m_+|d)$ during the embedding learning. 
To achieve this, we adopt the following margin-based ranking loss:
\begin{equation}
\begin{split}
    &\max \Big(0, \gamma + \log p(m_-|d) - \log p(m_+|d)\Big) \\
    \triangleq & \Big[\gamma + \log p(m_-|d) - \log p(m_+|d) \Big]_+.
\end{split}
\label{eqn:margin}
\end{equation}
Here, $m_-$ is a negative metadata context of document $d$; $\gamma > 0$ is a hyperparameter indicating the expected margin between a positive pair $(d, m_+)$ and a negative pair $(d, m_-)$. Based on the definition of $p(m|d)$ in Eq. (\ref{eqn:pv}), we have 
\begin{equation}
\begin{split}
    & \gamma + \log p(m_-|d) - \log p(m_+|d) \\
  =\ & \gamma + \log \frac{p(m_-|d)}{p(m_+|d)} \\
  =\ & \gamma +\log \frac{\exp(\bfe_{m_-}^T\bfe_d) / \big(\sum_{m' \in \mathcal{V}_m}\exp(\bfe_{m'}^T\bfe_d) \big)}{\exp(\bfe_{m_+}^T\bfe_d) / \big(\sum_{m' \in \mathcal{V}_m}\exp(\bfe_{m'}^T\bfe_d)\big)} \\
  =\ & \gamma + \log \frac{\exp(\bfe_{m_-}^T\bfe_d)}{\exp(\bfe_{m_+}^T\bfe_d)} \\
  =\ & \gamma + \bfe_{m_-}^T\bfe_d - \bfe_{m_+}^T\bfe_d.
\end{split}
\end{equation}
Therefore, the objective function of document-metadata proximity can be defined as follows.
\begin{equation}
\mathcal{J}_{\rm DM} = \sum_{d \in \mathcal{D}} \sum_{m_+ \in \mathcal{M}_d} \sum_{m_- \in \mathcal{V}_m \backslash \{m_+\}} \Big[ \gamma + \bfe_{m_-}^T\bfe_d - \bfe_{m_+}^T\bfe_d \Big]_+.
\end{equation}

\vspace{1mm}

\noindent \textbf{Document \& Label.} We have label information of each document in the training set. Therefore, the embedding pre-training step can be designed as a supervised process by incorporating those document-label relationships. 
Specifically, a document $d$ should be closer to its relevant labels $l_+$ than to its irrelevant labels $l_-$. To encourage this, we can define the conditional probability $p(l|d)$ in a form similar to Eq. (\ref{eqn:pv}). Then, following the derivation above, the objective of document-label proximity is
\begin{equation}
\mathcal{J}_{\rm DL} = \sum_{d \in \mathcal{D}}\sum_{l_+ \in \mathcal{L}_d} \sum_{l_- \in \mathcal{L}\backslash\{l_+\}} \Big[\gamma + \bfe_{l_-}^T\bfe_d - \bfe_{l_+}^T\bfe_d\Big]_+.
\end{equation}

\vspace{1mm}

\noindent \textbf{Document \& Word.} The document embedding $\bfe_d$ can be considered as the representation of the theme of $d$. Given a theme, authors write down words that are coherent with the meaning of the entire text. To encourage such coherence, we employ the following objective:
\begin{equation}
\mathcal{J}_{\rm DW} = \sum_{d \in \mathcal{D}}\sum_{w_+ \in \mathcal{W}_d} \sum_{w_- \in \mathcal{W}\backslash\{w_+\}} \Big[\gamma + \bfe_{w_-}^T\bfe_d - \bfe_{w_+}^T\bfe_d\Big]_+,
\end{equation}
where $\mathcal{W}_d$ is the text sequence of document $d$ and $\mathcal{W}$ is the whole word vocabulary.

\vspace{1mm}

\noindent \textbf{Word \& Context.} Given a text sequence $\mathcal{W}_d=w_1w_2\cdots w_N$, the semantic of a word $w_i$ depends on not only the document theme but also its surrounding words in the local context window $\mathcal{C}(w_i) = \{w_{i+j}|-x \leq j \leq x, j \neq 0\}$, where $x$ is the window size. Following \cite{mikolov2013distributed}, we assume each word has a center word embedding $\bfe_w$ and a context word embedding $\bfc_w$. To encourage the closeness between a word and its local context, the following objective can be proposed.
\begin{equation}
\mathcal{J}_{\rm WW} = \sum_{d \in \mathcal{D}}\sum_{w_+ \in \mathcal{W}_d} \sum_{w_- \in \mathcal{W}\backslash\{w_+\}} \sum_{w \in \mathcal{C}(w_+)} \Big[\gamma + \bfe_{w_-}^T\bfc_w - \bfe_{w_+}^T\bfc_w \Big]_+.
\end{equation}

Given the objective of each type of relationship, our embedding pre-training module can be formulated as a joint optimization problem as follows.
\begin{equation}
\begin{gathered}
\min_{\{\bfe_d\}, \{\bfe_m\}, \{\bfe_l\}, \{\bfe_w\}, \{\bfc_w\}} \mathcal{J}_{\rm embedding} = \mathcal{J}_{\rm DM} + \mathcal{J}_{\rm DL} + \mathcal{J}_{\rm DW} + \mathcal{J}_{\rm WW}, \\
\text{s.t.\ \ \ } ||\bfe_d||_2 = ||\bfe_m||_2 = ||\bfe_l||_2 = ||\bfe_w||_2 = ||\bfc_w||_2 = 1.
\end{gathered}
\end{equation}
We use the L2-norm constraints to control the scale of embedding vectors. These constraints are common when the margin-based ranking loss is used \cite{bordes2013translating,ren2017cotype}. Without these constraints, the gap between positive and negative pairs (e.g., $\bfe_{m_-}^T\bfe_d - \bfe_{m_+}^T\bfe_d$) can approach $-\infty$ when $||\bfe_d||_2$ becomes arbitrarily large, which makes the optimization problem trivial.

\vspace{1mm}

\noindent \textbf{Optimization.} The overall objective consists of four parts (i.e., $\mathcal{J}_{\rm DM}$, $\mathcal{J}_{\rm DL}$, $\mathcal{J}_{\rm DW}$ and $\mathcal{J}_{\rm WW}$). To optimize this objective, we adopt the sampling technique introduced in \cite{tang2015pte} for efficient updating. In each iteration, we alternatively optimize one part (e.g., $\mathcal{J}_{\rm DM}$) by randomly sampling a positive pair (e.g., $(d, m_+)$) and a corresponding negative pair (e.g., $(d, m_-)$). Given the two pairs, we can calculate the Euclidean gradient $\nabla^E$ of embeddings. Taking $\mathcal{J}_{\rm DM}$ as an example, the gradient vectors are as follows.
\begin{equation}
\begin{gathered}
\nabla^E \mathcal{J}_{\rm DM}(\bfe_d) = {\bf 1}(\gamma + \bfe_{m_-}^T\bfe_d - \bfe_{m_+}^T\bfe_d > 0)\cdot (\bfe_{m_-} - \bfe_{m_+}), \\
\nabla^E \mathcal{J}_{\rm DM}(\bfe_{m_+}) = {\bf 1}(\gamma + \bfe_{m_-}^T\bfe_d - \bfe_{m_+}^T\bfe_d > 0) \cdot (-\bfe_d), \\
\nabla^E \mathcal{J}_{\rm DM}(\bfe_{m_-}) = {\bf 1}(\gamma + \bfe_{m_-}^T\bfe_d - \bfe_{m_+}^T\bfe_d > 0) \cdot \bfe_d,
\end{gathered}
\end{equation}
where ${\bf 1}(\cdot)$ is the indicator function. When optimizing other parts, the Euclidean gradient can be calculated in a similar way.

Recall the constraints of our optimization problem that all embedding vectors need to reside on a sphere. Thus, Euclidean gradient approaches like SGD cannot be directly applied here. Instead, we adopt the Riemannian gradient method \cite{bonnabel2013stochastic}. Specifically, we calculate the Riemannian gradient $\nabla^R$ on a sphere based on the Euclidean gradient $\nabla^E$ according to the following equation \cite{meng2019spherical}:
\begin{equation}
\nabla^R \mathcal{J}(\bfe) = ({\bm I}-\bfe\bfe^T)\nabla^E \mathcal{J}(\bfe).
\end{equation}
Then we update the embedding vectors in the following form \cite{bonnabel2013stochastic}:
\begin{equation}
\bfe^{(t+1)} \leftarrow \frac{\bfe^{(t)}+\alpha_t \nabla^R \mathcal{J}(\bfe^{(t)})}{||\bfe^{(t)}+\alpha_t \nabla^R \mathcal{J}(\bfe^{(t)})||_2},
\end{equation}
where $\alpha_t$ is the learning rate at step $t$.

There are several other ways to jointly embed heterogeneous signals \cite{tang2015pte,dong2017metapath2vec}. For example, PTE \cite{tang2015pte} constructs three bipartite graphs describing the relationships between labels, words and documents and then embeds these elements into the same latent space. We would like to mention two key differences between our pre-training step and PTE: First, we propose to use a margin-based ranking loss with metadata instances included as well. Second, we formulate the optimization problem in a spherical space and solve it by using the Riemannian gradient method.

\subsection{Transformer Layers}
\label{sec:trans}
Given a document, to facilitate extensive information exchange between text and metadata during document encoding, we adopt the Transformer architecture \cite{vaswani2017attention} as our encoder. Transformer proposes a fully connected attention mechanism to support such exchange between any two tokens in a sequence. 
Therefore, we concatenate all metadata instances of a document with its word sequence to form the layer input. 
Moreover, we add [CLS] tokens at the beginning of each input sequence. First proposed in BERT \cite{devlin2019bert}, the final state of such special tokens are used as aggregate sequence representation for classification tasks. When the label space is large (e.g., 10K), one [CLS] token (e.g., a 100-dimensional vector) may not be informative enough to predict the relevant labels. Therefore, following \cite{xun2020correlation}, we put multiple [CLS] tokens [CLS$_1$], ..., [CLS$_C$] in the input.
To summarize, given a document $d$, the layer input $\bfH$ is
\begin{equation}
    \bfH = \big[ \underbrace{\vphantom{\bfe_{p_1}} \bfe_{[{\rm CLS}_1]}; ...; \bfe_{[{\rm CLS}_C]}}_{{\rm [CLS]\ tokens}}; \ \  \underbrace{\vphantom{\bfe_{p_1}} \bfe_{m_1}; ...; \bfe_{m_M}}_{{\rm metadata}\ \mathcal{M}_d}; \ \  \underbrace{\vphantom{\bfe_{p_1}} \bfe_{w_1}; ...; \bfe_{w_N}}_{{\rm words}\ \mathcal{W}_d}  \big]. \notag
\end{equation}
Here, $\bfH \in \mathbb{R}^{\delta \times (C+|\mathcal{M}_d|+|\mathcal{W}_d|)}$, where $\delta$ is the dimension of the embedding space.

\begin{example}
\textsc{(Input Sequence)} \textit{Suppose we are given the document ``Graph structure in the Web'' in Figure \ref{fig:framework}. The input sequence of the Transformer layer will be}

\vspace{1mm}

\noindent `` {\color{darkg} [CLS$_1$]\ \ ... \ \ [CLS$_C$]}\ \ {\color{blue} [\textsc{Venue}\_WWW]\ \ [\textsc{Author}\_Andrei Broder] [\textsc{Author}\_Ravi Kumar] ... [\textsc{Reference}\_2066636486] [\textsc{Reference} \_1976969221]\ \ ...}\ \ {\color{orange} [\textsc{Word}\_graph]\ \ [\textsc{Word}\_structure]\ \ [\textsc{Word}\_in] [\textsc{Word}\_the] [\textsc{Word}\_web] ...} ''

\vspace{1mm}

\noindent \textit{Here, the green tokens represent [CLS] symbols; the blue tokens denote metadata instances (i.e., venue, authors and references in this specific example); the orange tokens represent words in the document.}
\end{example}

\noindent \textbf{Intuition behind the Metadata-aware Input Sequence.} Previous studies (e.g., \cite{guo2019star}) have pointed out that, given an input sequence $\mathcal{S}$, Transformer treats $\mathcal{S}$ as a fully connected token graph. For each token $i \in \mathcal{S}$, its context is the entire sequence, and its representation will be updated by aggregating the information from all tokens $j \in \mathcal{S}$.
In our case, $\mathcal{S}$ is the union of $\mathcal{M}_d$, $\mathcal{W}_d$ and [CLS] tokens. Hence, the attention mechanism allows each [CLS] token to aggregate information from all metadata instances and words. Moreover, if we treat each input document as an ego network of the document node $d$ (as shown in Figure \ref{fig:framework}), our embedding pre-training step essentially captures first-order proximity between $d$ and its neighbors, while the fully connected attention mechanism here describes second-order proximity in $d$'s neighborhood. In other words, our Transformer layer facilitates higher-order interactions among metadata instances and words.

\vspace{1mm}

\noindent \textbf{Multi-head Attention.} Now we formally introduce the attention mechanism in the Transformer layer. As in \cite{vaswani2017attention}, given $\bfH$, one can use a query vector $\bfq \in \mathbb{R}^{1 \times \delta}$ to select relevant information with attention.
\begin{equation}
    \textrm{Attention}(\bfq, \bfK, \bfV) = \textrm{Softmax}\Big(\frac{\bfq\bfK^T}{\sqrt{\delta}}\Big)\bfV,
\end{equation}
where $\bfK = \bfH \bfW^K$ and $\bfV = \bfH \bfW^V$. Matrices $\bfW^K$ and $\bfW^V$ are parameters to be learned.

Similar to the idea of multiple channels in CNN, Transformer uses multi-head attention to extract more signals from $\bfH$. Formally,
\begin{equation}
    \begin{gathered}
    \bfa_i = \textrm{Attention}(\bfq\bfW^Q_i, \bfH\bfW^K_i, \bfH\bfW^V_i), \\
    \textrm{MultiHeadAtt}(\bfq, \bfH) = \big[\bfa_1\ ||\ \bfa_2\ ||\ ...\ ||\ \bfa_k\big]\bfW^O,
    \end{gathered}
\end{equation}
where $||$ denotes the concatenation operation. Matrices $\bfW^Q_i$, $\bfW^K_i$, $\bfW^V_i$ and $\bfW^O$ are learnable parameters.

\vspace{1mm}

\noindent \textbf{Document Encoding.} Using multi-head attention, for each input token $i \in \bfH$, we update its representation based on its pre-trained embedding $\bfe_i$.
\begin{equation}
    \begin{gathered}
    \bfz_i = \textrm{LayerNorm}\big(\bfe_i+\textrm{MultiHeadAtt}(\bfe_i, \bfH)\big), \\
    \bfh_i = \textrm{LayerNorm}\big(\bfz_i + \textrm{FFN}(\bfz_i)\big).
    \end{gathered}
\label{eqn:layer}
\end{equation}
Here, $\textrm{LayerNorm}(\cdot)$ is the layer normalization operator \cite{ba2016layer} and $\textrm{FFN}(\cdot)$ is the position-wise feed-forward network \cite{vaswani2017attention}. 
To incorporate position information of the token, we further concatenate its sinusoidal position embedding \cite{vaswani2017attention} with its input embedding $\bfe_i$.

Eq. (\ref{eqn:layer}) describes one Transformer layer. As shown in Figure \ref{fig:framework}, we can stack $L$ Transformer layers, where the output of the $l$-th layer $\bfH^{(l)}$ is also the input of the $(l+1)$-th layer. $\bfH^{(0)}$ consists of the pre-trained embeddings, and $\bfH^{(L)}$ is used for prediction.

\subsection{Prediction Layer}
\label{sec:pred}
After $L$ Transformer layers, we concatenate the final state of all [CLS] tokens to get the final document representation $\widehat{\bfh}_d$.
\begin{equation}
\widehat{\bfh}_d = \bfh^{(L)}_{[{\rm CLS}_1]}\ ||\ \bfh^{(L)}_{[{\rm CLS}_2]}\ ||\ ...\ ||\ \bfh^{(L)}_{[{\rm CLS}_C]}.
\end{equation}

To perform classification, we add a fully connected layer upon the output of Transformer. 
The final layer is then connected to $|\mathcal{L}|$ sigmoid functions, which correspond to all labels in $\mathcal{L}$. The output of the $l$-th sigmoid function ($\uppi_{dl}$) denotes the probability that document $d$ should be tagged with label $l$. Formally,
\begin{equation}
    \bfpi_d = \textrm{Sigmoid}\big(\widehat{\bfh}_d\bfW^{\Pi}+\bfb\big),
\end{equation}
where $\bfW^{\Pi} = [\bfw_1,...,\bfw_{|\mathcal{L}|}]$ and $\bfw_l$ can be viewed as the parameters specific to the $l$-th label.

Given the output probabilities, our model minimizes the binary cross-entropy (BCE) loss by treating the multi-label classification task as $|\mathcal{L}|$ binary classification subtasks.
\begin{equation}
    \mathcal{J}_{\rm BCE} = -\sum_{d\in \mathcal{D}}\sum_{l\in \mathcal{L}} \big(y_{dl}\log\uppi_{dl} + (1-y_{dl})\log(1-\uppi_{dl}) \big),
\end{equation}
where $y_{dl}=1$ means document $d$ has label $l$, and $y_{dl}=0$ otherwise.

\subsection{Hypernymy Regularization}
In hierarchical text classification, a given label taxonomy contains valuable signals of label intercorrelation, which should be leveraged in the classification process. 
However, most existing studies ignore the label dependencies in the input taxonomy \cite{liu2017deep,xun2019meshprobenet,you2019attentionxml}.  

To incorporate the label hierarchy into \textsf{\model}, we propose to regularize each non-root label by its parents. 
Specifically, the regularization is applied in both the parameter space and the output space. 
In the parameter space, instead of treating the class-specific parameters $\bfw_1,...,\bfw_{|\mathcal{L}|}$ as independent, we design a regularization mechanism for modeling the dependencies in the prediction layer; 
In the output space, we enable the interactions between the output probabilities $\uppi_{d1},...,\uppi_{d|\mathcal{L}|}$  in the loss function.

\vspace{1mm}

\noindent \textbf{Regularization in the Parameter Space.} Similar to \cite{gopal2013recursive,peng2018large}, we use an L2-norm penalty to enforce the parameters of each label to be similar to its parent. 
\begin{equation}
    \mathcal{J}_{\rm parameter} = \sum_{l\in \mathcal{L}} \sum_{l' \in \Phi(l)} \frac{1}{2}||\bfw_l - \bfw_{l'}||^2,
\label{eqn:parareg}
\end{equation}
where $\Phi(l)$ denotes the set of parent labels of $l$. Intuitively, this regularization encourages comparable criteria of categories that are nearby in the hierarchy. For example, judging whether a document can be tagged with ``\texttt{Crawling}'' should bear similarities with judging whether it is related to its parent label ``\texttt{World Wide Web}''.

\vspace{1mm}

\noindent \textbf{Regularization in the Output Space.} Previous studies on hierarchical regularization \cite{gopal2013recursive,peng2018large} only consider the ``similarity'' between parent and child labels. To be specific, in Eq. (\ref{eqn:parareg}), the L2-norm is symmetric on the child $l$ and the parent $l'$. In other words, even if we swap $l$ and $l'$, the regularization term for $\bfw_l$ and $\bfw_{l'}$ remains unchanged. This could be insufficient to capture the asymmetry between parent and child labels. To address this issue, inspired by the distributional inclusion hypothesis (DIH) \cite{geffet2005distributional}, we propose a novel regularization term to characterize the hypernym-hyponym relationships.

\begin{definition}
\textsc{(Distributional Inclusion Hypothesis \cite{geffet2005distributional})} \textit{If the meaning of a word $w_1$ entails another word $w_2$, then it is expected that all the typical contexts of $w_1$ will also occur with $w_2$.}
\end{definition}
According to this definition, $w_2$ is viewed as a hypernym (i.e., parent) and $w_1$ is viewed as a hyponym (i.e., child). Note that one can interpret DIH in various ways depending on how ``contexts'' are defined. For example, if ``contexts'' are defined as documents \cite{shi2019discovering}, then DIH states that: if a word (e.g., ``\texttt{Crawling}'') appears in a document, then its parent (e.g., ``\texttt{World Wide Web}'') is also expected to be in that document. In contrast, if ``contexts'' are defined based on the local context window (i.e., the previous and the latter words in a sequence) \cite{shwartz2017hypernyms}, then DIH becomes: if a context word $c$ occurs $n$ times in the context window of a child $w_1$, then it is expected to occur no less than $n$ times in the context window of its parent $w_2$. 
DIH is a classic tool in constructing topic taxonomies \cite{shi2019discovering,shen2018web},
which motivates us to propose the following DIH-based regularization.

In the document classification task, the hypernym $w_2$ and hyponym $w_1$ become the parent label $l'$ and child label $l$, respectively. We define the ``contexts'' of a label $l$ to be the documents tagged with $l$. From this perspective, DIH can be interpreted as: if a document $d$ belongs to the child class $l$ with probability $\uppi_{dl}$, then it should belong to the parent class $l'$ with probability no less than $\uppi_{dl}$. For example, if there is a 50\% chance a paper will be labeled with ``\texttt{Crawling}'', then the chance to tag this paper with ``\texttt{World Wide Web}'' should be at least 50\%. Formally, the regularization term is defined as
\begin{equation}
    \mathcal{J}_{\rm output} = \sum_{d\in \mathcal{D}} \sum_{l\in \mathcal{L}} \sum_{l' \in \Phi(l)} \max\big(0, \uppi_{dl} - \uppi_{dl'}\big).
\label{eqn:outreg}
\end{equation}
Unlike the parameter regularization, Eq. (\ref{eqn:outreg}) is asymmetric: $\uppi_{dl} > \uppi_{dl'}$ will incur a penalty, but $\uppi_{dl'} > \uppi_{dl}$ will not. 

Based on the BCE loss and the two proposed regularization terms, we use the following objective to learn the parameters of our neural architecture:
\begin{equation}
   \min \mathcal{J} = \mathcal{J}_{\rm BCE} + \lambda_1 \mathcal{J}_{\rm parameter} + \lambda_2 \mathcal{J}_{\rm output},
\label{eqn:obj}
\end{equation}
where $\lambda_1$ and $\lambda_2$ are two hyperparameters.

\section{Experiments}
\subsection{Setup}
\noindent \textbf{Datasets.} We evaluate our method on two large-scale datasets.
\begin{itemize}[leftmargin=*]
    \item \textbf{MAG-CS \cite{wang2020microsoft}.} The Microsoft Academic Graph (MAG) has a web-scale collection of scientific papers covering a broad spectrum of academic disciplines. As of February 2021, it has more than 251 million academic papers and over 729 thousand labels. MAG has also performed author name disambiguation and represented each author with a unique ID. Based on MAG, we construct a dataset focusing on the computer science domain. Specifically, we select papers published at 105 top CS conferences\footnote{\url{https://github.com/microsoft/mag-covid19-research-examples/blob/master/src/MAG-Samples/impact-of-covid19-on-the-computer-science-research-community/TopCSConferences.txt}} from 1990 to 2020. MAG has a high-quality label taxonomy constructed semi-automatically \cite{shen2018web}. For each selected paper, we remove its labels that are not in the CS domain (i.e., not descendants of ``\texttt{Computer Science}'' in the taxonomy). We also remove the root label ``\texttt{Computer Science}'' which is trivial to predict. After paper selection and label filtering, we obtain 705,407 documents and 15,809 labels. We refer to this dataset as MAG-CS.
    \item \textbf{PubMed \cite{lu2011pubmed}.} PubMed comprises more than 30 million articles (abstracts) of biomedical literature from MEDLINE, life science journals, and online books. In our experiment, we focus on papers published in 150 top journals in medicine\footnote{\url{https://academic.microsoft.com/journals/71924100}} from 2010 to 2020. For each paper selected from PubMed, we find it in MAG so that we can obtain its disambiguated author, venue, and reference information. Each PubMed paper is tagged with related MeSH terms \cite{coletti2001medical}, which are viewed as labels in our task. In the MeSH hierarchy, we focus on the first 8 top-level categories (i.e., A--H)\footnote{\url{https://meshb.nlm.nih.gov/treeView}}. After selection, we have 898,546 documents and 17,693 labels.
\end{itemize}

For both datasets, we use 80\% of the documents for training, 10\% for validation, and 10\% for testing. 
The text information of each document is its title and abstract; the metadata information includes authors, venue, and references. Table \ref{tab:stat} summarizes the statistics of the two datasets.

\newcolumntype{C}[1]{>{\centering\let\newline\\\arraybackslash\hspace{0pt}}m{#1}}
\begin{table}[t]
\centering
\caption{Dataset statistics.}
\vspace{-0.5em}
\small
\begin{tabular}{C{2.9cm} C{2.0cm} C{2.0cm}}
\hline
                      & \textbf{MAG-CS \cite{sinha2015overview}}  & \textbf{PubMed \cite{lu2011pubmed}}  \\ \hline
\# Training Docs      & 564,340 & 718,837 \\
\# Validation Docs    & 70,534  & 89,855  \\
\# Testing Docs       & 70,533  & 89,854  \\ \hline
\# Labels             & 15,809  & 17,963  \\
\# Labels / Doc       & 5.60    & 7.78    \\ \hline
Vocabulary Size       & 425,316  & 776,975  \\
\# Words / Doc        & 126.33  & 198.97   \\ \hline
\# Authors            & 818,927  & 2,201,919  \\
\# Venues             & 105  & 150  \\
\# Paper-Author Edges & 2,274,546    & 5,989,142 \\ 
\# Paper-Venue Edges  & 705,407  &  898,546 \\
\# Paper-Paper Edges  & 1,518,466     & 4,455,702 \\ \hline
\# Edges in Taxonomy  & 27,288  & 22,842  \\
\# Layers of Taxonomy & 6       & 15      \\   \hline
\end{tabular}
\label{tab:stat}
\vspace{-0.5em}
\end{table}

\begin{table*}[t]
\caption{Performance of compared algorithms on MAG-CS. *: significantly worse than \textsf{\model} (p-value $< 0.05$). **: significantly worse than \textsf{\model} (p-value $< 0.01$).}
\vspace{-0.5em}
\small
\begin{tabular}{c C{2.4cm} C{2.35cm} C{2.35cm} C{2.35cm} C{2.35cm}}
\hline
\textbf{Algorithms}       & \textbf{P@1$=$NDCG@1} & \textbf{P@3} & \textbf{P@5} & \textbf{NDCG@3} & \textbf{NDCG@5} \\ \hline
XML-CNN \cite{liu2017deep}                  &  0.8656 $\pm$ 0.0006**          &  0.7028 $\pm$ 0.0010**   & 0.5756 $\pm$ 0.0010**    &  0.7842 $\pm$ 0.0009**      &  0.7407 $\pm$ 0.0009**      \\
MeSHProbeNet \cite{xun2019meshprobenet}     &  0.8738 $\pm$ 0.0016**          & 0.7219 $\pm$ 0.0059**    & 0.5927 $\pm$ 0.0075**    &  0.8020 $\pm$ 0.0048**      &   0.7588 $\pm$ 0.0067**     \\
AttentionXML \cite{you2019attentionxml}     &  0.9035 $\pm$ 0.0009**          & 0.7682 $\pm$ 0.0017**    & 0.6441 $\pm$ 0.0020    &  0.8489 $\pm$ 0.0016**      &   0.8145 $\pm$ 0.0020**     \\ \hline
Star-Transformer \cite{guo2019star}         &  0.8569 $\pm$ 0.0011**          & 0.7089 $\pm$ 0.0010**    & 0.5853 $\pm$ 0.0011**    &  0.7876 $\pm$ 0.0008**      &   0.7486 $\pm$ 0.0011**     \\
BERTXML \cite{xun2020correlation}           & 0.9011 $\pm$ 0.0027**           &  0.7532 $\pm$ 0.0015**   & 0.6238 $\pm$ 0.0020*    & 0.8355 $\pm$ 0.0025**       &   0.7954 $\pm$ 0.0024**     \\
Transformer \cite{vaswani2017attention}     &  0.8805 $\pm$ 0.0007**          & 0.7327 $\pm$ 0.0006**    & 0.6024 $\pm$ 0.0010**    &  0.8129 $\pm$ 0.0008**      &   0.7703 $\pm$ 0.0010**     \\ \hline
\textsf{\model}-NoMetadata                    &  0.9041 $\pm$ 0.0012**          & 0.7640 $\pm$ 0.0010*    & 0.6376 $\pm$ 0.0002*    &  0.8440 $\pm$ 0.0012**      &   0.8068 $\pm$ 0.0005**     \\
\textsf{\model}-NoHierarchy                    &  0.9114 $\pm$ 0.0014*          & 0.7634 $\pm$ 0.0012**    & 0.6312 $\pm$ 0.0013**    &  0.8486 $\pm$ 0.0006**      &   0.8076 $\pm$ 0.0009**     \\
\textsf{\model}                               &  \textbf{0.9190 $\pm$ 0.0012}          & \textbf{0.7763 $\pm$ 0.0023}    & \textbf{0.6457 $\pm$ 0.0030}    &  \textbf{0.8610 $\pm$ 0.0022}      &   \textbf{0.8223 $\pm$ 0.0030}     \\ \hline
\end{tabular}
\label{tab:mag}
\end{table*}

\begin{table*}[t]
\caption{Performance of compared algorithms on PubMed. *: significantly worse than \textsf{\model} (p-value $< 0.05$). **: significantly worse than \textsf{\model} (p-value $< 0.01$).}
\vspace{-0.5em}
\small
\begin{tabular}{c C{2.4cm} C{2.35cm} C{2.35cm} C{2.35cm} C{2.35cm}}
\hline
\textbf{Algorithms}       & \textbf{P@1$=$NDCG@1} & \textbf{P@3} & \textbf{P@5} & \textbf{NDCG@3} & \textbf{NDCG@5} \\ \hline
XML-CNN \cite{liu2017deep}                  &  0.9084 $\pm$ 0.0004**          &  0.7182 $\pm$ 0.0007**   & 0.5857 $\pm$ 0.0004**    &  0.7790 $\pm$ 0.0007**      &  0.7075 $\pm$ 0.0005**      \\
MeSHProbeNet \cite{xun2019meshprobenet}     &  0.9135 $\pm$ 0.0021          & 0.7224 $\pm$ 0.0066*    & 0.5878 $\pm$ 0.0070*    &  0.7836 $\pm$ 0.0057*      &   0.7109 $\pm$ 0.0065*     \\
AttentionXML \cite{you2019attentionxml}     &  0.9125 $\pm$ 0.0003*          & 0.7414 $\pm$ 0.0017*    & 0.6169 $\pm$ 0.0016    &  0.7979 $\pm$ 0.0013*      &   0.7341 $\pm$ 0.0013     \\ \hline
Star-Transformer \cite{guo2019star}         &  0.8962 $\pm$ 0.0023**          & 0.6990 $\pm$ 0.0014**    & 0.5641 $\pm$ 0.0008**    &  0.7612 $\pm$ 0.0015**      &   0.6869 $\pm$ 0.0011**     \\
BERTXML \cite{xun2020correlation}           & 0.9144 $\pm$ 0.0014*           &  0.7362 $\pm$ 0.0046*   & 0.6032 $\pm$ 0.0050*    & 0.7949 $\pm$ 0.0038*       &   0.7247 $\pm$ 0.0045*     \\
Transformer \cite{vaswani2017attention}     &  0.8971 $\pm$ 0.0050*          & 0.7299 $\pm$ 0.0029**    & 0.6003 $\pm$ 0.0018**    &  0.7867 $\pm$ 0.0034**      &   0.7178 $\pm$ 0.0027**     \\ \hline
\textsf{\model}-NoMetadata                    &  0.9153 $\pm$ 0.0022          & 0.7408 $\pm$ 0.0035*    & 0.6080 $\pm$ 0.0036**    &  0.7987 $\pm$ 0.0031*      &   0.7290 $\pm$ 0.0034*     \\
\textsf{\model}-NoHierarchy                    &  0.9151 $\pm$ 0.0022          & 0.7425 $\pm$ 0.0041    & 0.6104 $\pm$ 0.0047    &  0.8001 $\pm$ 0.0037      &   0.7310 $\pm$ 0.0044     \\
\textsf{\model}                               &  \textbf{0.9168 $\pm$ 0.0013}          & \textbf{0.7511 $\pm$ 0.0029}    & \textbf{0.6199 $\pm$ 0.0029}    &  \textbf{0.8072 $\pm$ 0.0027}      &   \textbf{0.7395 $\pm$ 0.0029}     \\ \hline
\end{tabular}
\label{tab:mesh}
\end{table*}

\vspace{1mm}

\noindent \textbf{Compared Methods.}
We compare the following approaches including both extreme multi-label text classification methods as well as Transformer-based models.
\begin{itemize}[leftmargin=*]
    \item \textbf{XML-CNN \cite{liu2017deep}} is an extreme multi-label text classification method based on convolutional neural networks. It modifies Kim-CNN \cite{kim2014convolutional} by introducing a dynamic max-pooling scheme, a bottleneck layer, and the BCE loss.
    \item \textbf{MeSHProbeNet \cite{xun2019meshprobenet}} was originally designed for tagging biomedical documents with relevant MeSH terms. It can also be applied to a general multi-label text classification setting.
    MeSHProbeNet models text sequences using recurrent neural networks and uses multiple MeSH ``probes'' to extract information from RNN hidden states.
    \item \textbf{AttentionXML \cite{you2019attentionxml}} is an extreme multi-label text classification method built upon a bidirectional RNN layer and a label-aware attention layer. It also leverages hierarchical label trees to recursively warm-start the model.
    \item \textbf{Transformer \cite{vaswani2017attention}} is a fully connected attention-based model. Since we have massive training data in both datasets, we train a Transformer encoder from scratch using text classification as the downstream task. Following \cite{liu2019neuralclassifier}, after getting the output representation of all tokens, we average them to get document representation and pass it through a fully connected layer to perform multi-label classification.
    \item \textbf{Star-Transformer \cite{guo2019star}} simplifies Transformer by sparsifying fully connected attention to a star-shaped structure. This sparsification leads to performance improvement on moderately sized training sets. 
    \item \textbf{BERTXML \cite{xun2020correlation}} is a model inspired by BERT \cite{devlin2019bert}. It utilizes a multi-layer Transformer structure and adds multiple [CLS] symbols in front of the input sequence to obtain the aggregate sequence representation.
    \item \textsf{\textbf{\model}} is our proposed model with metadata-aware pre-training, metadata-aware Transformer encoding, and hypernymy regularization.
    \item \textsf{\textbf{\model}}\textbf{-NoMetadata} is an ablation version of the full \textsf{\model} model without using metadata information in both pre-training and Transformer layers.
    \item \textsf{\textbf{\model}}\textbf{-NoHierarchy} is an ablation version of the full \textsf{\model} model without hypernymy regularization.
\end{itemize}

\noindent \textbf{Implementation and Hyperparameters.} For all compared algorithms, the embedding dimension $\delta$ is 100. 
We use GloVe.6B.100d \cite{pennington2014glove} as initialized word embeddings for all models except \textsf{\model} and \textsf{\model}-NoHierarchy (whose initialized embeddings are learned from metadata-aware pre-training).
The training process is performed using Adam \cite{kingma2014adam} with a batch size of 256.
The baselines are implemented in two GitHub repositories\footnote{https://github.com/XunGuangxu/CorNet} \footnote{\url{https://github.com/Tencent/NeuralNLP-NeuralClassifier}}.
We directly use their default parameter settings when running the baselines.

For our \textsf{\model} framework, we set the margin of embedding pre-training $\gamma=0.3$, number of attention heads $k=2$, number of [CLS] tokens $C=8$, number of Transformer layers $L=3$, and the dropout rate to be 0.1.

\vspace{1mm}

\begin{figure*}
\centering
\subfigure[MAG-CS]{
\includegraphics[width=0.43\textwidth]{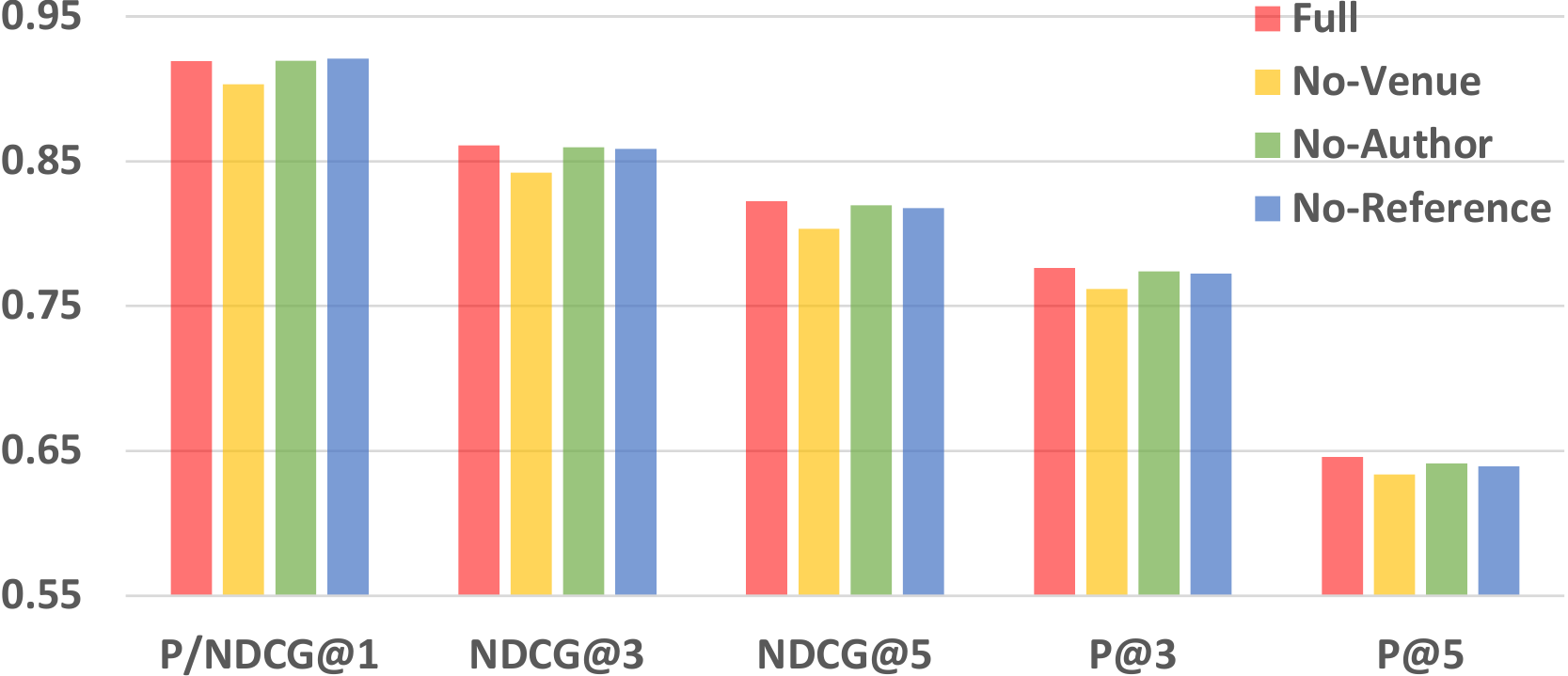}}
\hspace{2.5em}
\subfigure[PubMed]{
\includegraphics[width=0.43\textwidth]{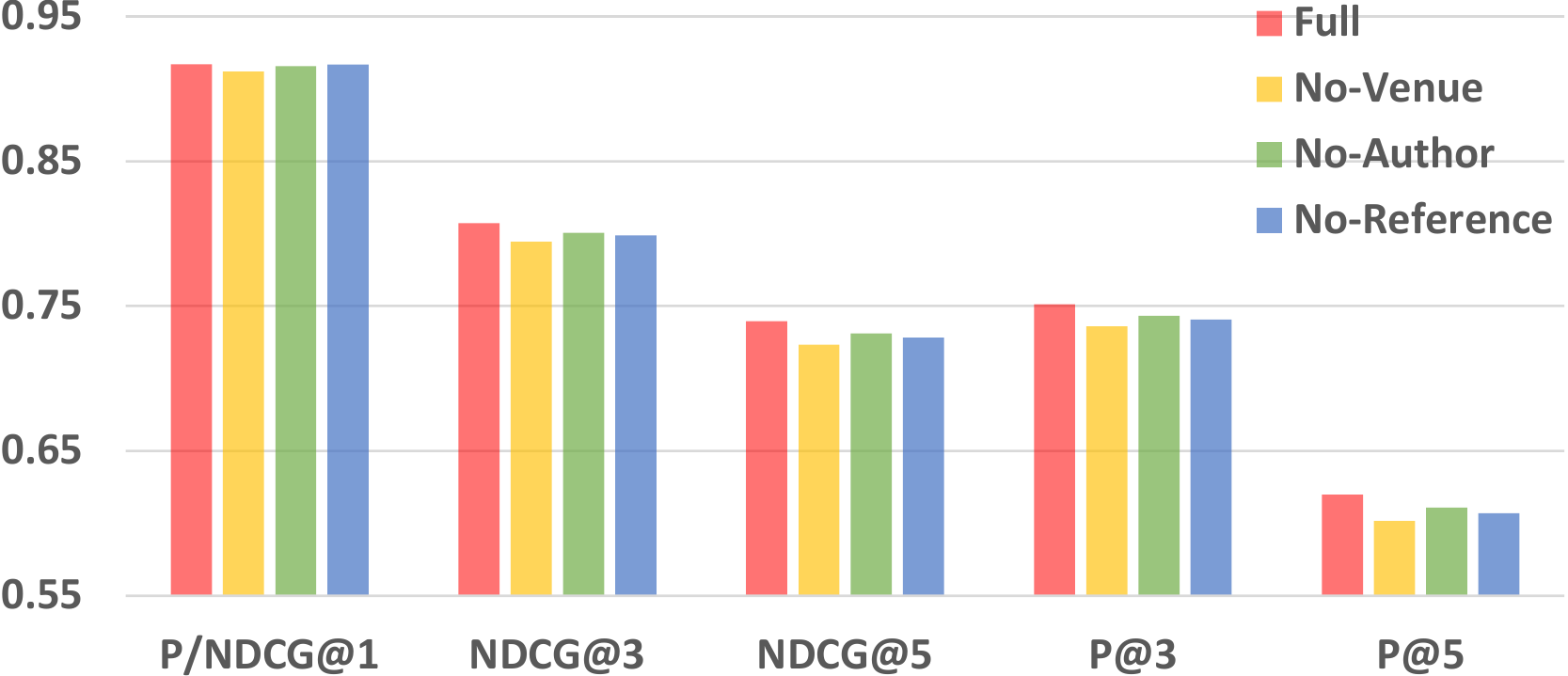}}
\vspace{-1em}
\caption{Ablation analysis of metadata.} 
\vspace{-0.5em}
\label{fig:metadata}
\end{figure*}

\noindent \textbf{Evaluation Metrics.} In many multi-label classification datasets, even if the label space is large, each document only has very few relevant labels. For example, in Table 1, we show that both MAG-CS and PubMed have over 15K labels in total, but each document has 5.60 and 7.78 labels on average, respectively. Considering the sparsity of labels, a short-ranked list of potentially relevant labels for each testing document is commonly used to represent classification quality. Following previous studies on extreme multi-label text classification \cite{liu2017deep,you2019attentionxml,xun2020correlation}, we adopt two rank-based metrics: the precision at top $k$ (P@$k$) and the normalized Discounted Cumulative Gain at top $k$ (NDCG@$k$), where $k=1,3,5$. For a document $d$, let $\bfy_d \in \{0,1\}^{|\mathcal{L}|}$ be its ground truth label vector and ${\rm rank}(i)$ be the index of the $i$-th highest predicted label according to the output probability $\bfpi_d$. Then, P@$k$ and NDCG@$k$ are formally defined as 
\begin{equation}
    \begin{gathered}
    {\rm P@}k = \frac{1}{k} \sum_{i = 1}^k y_{d, {\rm rank}(i)}. \\
    {\rm DCG@}k = \sum_{i=1}^k \frac{y_{d, {\rm rank}(i)}}{\log(i+1)}, \\
    {\rm NDCG@}k = \frac{{\rm DCG@}k}{\sum_{i=1}^{\min(k, ||\bfy_d||_0)}\frac{1}{\log(i+1)}}.
    \end{gathered}
\end{equation}
It is easy to show that ${\rm P@}1 \equiv {\rm NDCG@}1$ if each document has at least one true label.

\subsection{Performance Comparison}
Tables \ref{tab:mag} and \ref{tab:mesh} demonstrate the performance of compared algorithms on MAG-CS and PubMed, respectively. We run each experiment three times with the mean and standard deviation reported. To measure statistical significance, we conduct a two-tailed paired t-test to compare \textsf{\model} and each baseline. The significance level of each result is marked in the tables.

On MAG-CS, as we can observe from Table \ref{tab:mag}: (1) \textsf{\model} consistently outperforms all baseline approaches. In almost all cases, the gap is statistically significant, with only one exception where P@5 of AttentionXML is close to that of \textsf{\model}. 
(2) \textsf{\model} also significantly outperforms the two ablation versions \textsf{\model}-NoMetadata and \textsf{\model}-NoHierarchy. This observation validates our claim that both metadata and hierarchy signals are beneficial to the classification performance.
(3) Although Star-Transformer is shown to be more effective and efficient than the standard Transformer for modestly sized training sets \cite{guo2019star}, its simplified structure is less capable of fitting large-scale training sets. The comparison between Star-Transformer and Transformer in Table \ref{tab:mag} shows that MAG-CS is large enough to train a fully connected Transformer architecture from scratch. 
(4) The standard Transformer outperforms two dedicated multi-label text classification approaches, XML-CNN and MeSHProbeNet, which demonstrates the advantage of Transformer's fully connected attention mechanism over CNN and RNN architectures on MAG-CS. Built upon Transformer, \textsf{\model} can also outperform XML-CNN and MeSHProbeNet, even without metadata information. 

On PubMed, \textsf{\model} still performs the best among all compared approaches, and most observations from Table \ref{tab:mag} hold in Table \ref{tab:mesh}. However, we would like to emphasize one unique finding: 
the contribution of hypernymy regularization is no longer significant on PubMed. To be specific, on MAG-CS, \textsf{\model} has an average \textit{absolute} improvement of 1.2\% on the five metrics in comparison with \textsf{\model}-NoHierarchy; on PubMed, the improvement becomes 0.7\%. We believe this is due to different labeling patterns on the two datasets. As we can see, the effect of hypernymy regularization depends on the correlation between parent and child labels. In fact, when a document is tagged with a child label, we expect it will be labeled with its parents as well. However, this assumption is not often correct on PubMed as sometimes human annotators will only select those more specific categories to annotate the document. On MAG-CS, the assumption holds in more cases because each document is guaranteed to have at least one layer-1 label.

\subsection{Effect of Metadata}
In both datasets, we have three types of metadata information: authors, venue, and references. To check whether each of them is useful, we conduct an ablation analysis to study the performance change when \textsf{\model} is blind to one type of metadata. To do this, we create three ablation versions of \textsf{\model}: \textbf{No-Author}, \textbf{No-Venue}, and \textbf{No-Reference}. For No-Author, we remove author information from the input metadata $\mathcal{M}_d$ of each document $d$. Similarly, we can define No-Venue and No-Reference.

Figure \ref{fig:metadata} depicts the comparisons between \textsf{\model} and its three ablations. We observe that: (1) The full \textsf{\model} model outperforms No-Author, No-Venue, and No-Reference in most cases, indicating that all three types of metadata play a positive role in the classification process. (2) Among the three ablation versions, No-Venue consistently performs the worst. In other words, venue information has the largest contribution. In fact, when $k=1$, the contribution of authors and references to P/NDCG@$k$ is quite subtle, while venue signals have an evident offering. To explain this, we recall the hypernymy regularization inspired by DIH. We expect the predicted probability of a parent category to be no less than that of its children. Thus, the more general a label is, the higher probability it is expected to have. That being said, layer-1 categories are assumed to be ranked higher in the prediction list. Therefore, as strong indicators of coarse-grained classes (e.g., ``\texttt{Data Mining}'' and ``\texttt{Natural Language Processing}''), venues are expected to be most helpful to predict the higher-ranked labels. Since venues already give enough hints, overlooking authors or references will not lead to a visible performance drop when $k=1$. (3) As $k$ increases, the contribution of authors and references becomes larger. For example, on PubMed, the difference of P@1 between Full and No-Author is 0.1\%, but the difference of P@5 becomes 0.9\%. This is because venues are less beneficial to the prediction of fine-grained categories (e.g., ``\texttt{Named Entity Recognition}'' and ``\texttt{Entity Linking}''), but authors and references may provide such signals.

\subsection{Effect of Embedding Pre-Training}
\begin{figure}
\centering
\subfigure[MAG-CS, Training Loss]{
\includegraphics[width=0.23\textwidth]{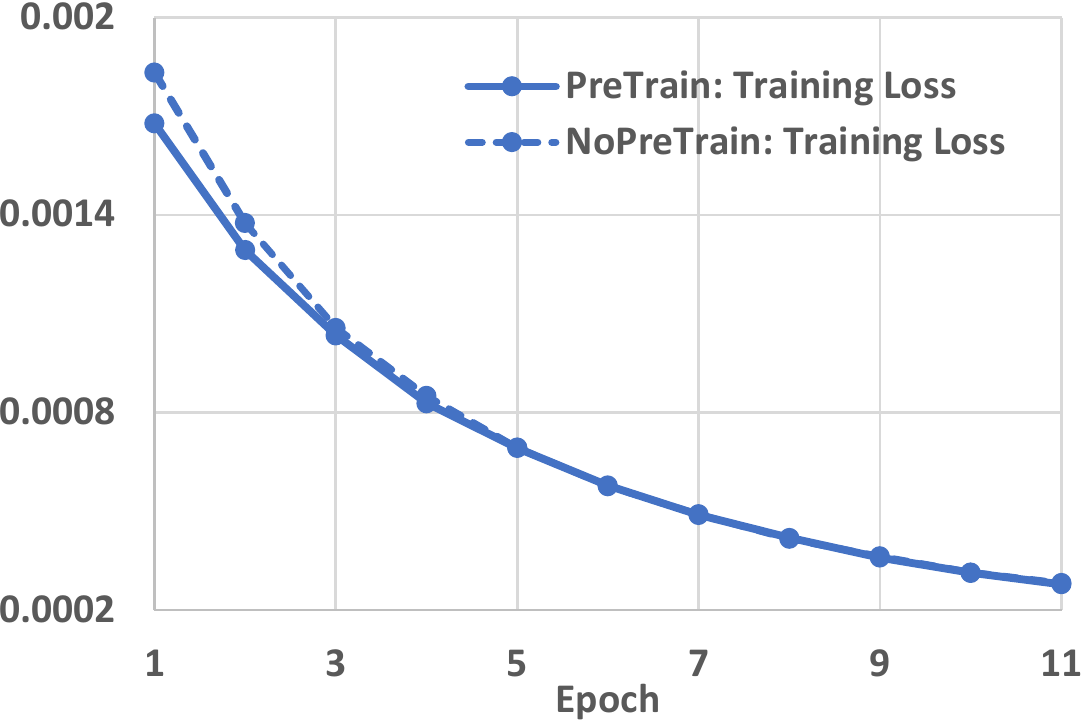}}
\hspace{-0.25em}
\subfigure[MAG-CS, Validation NDCG@$k$]{
\includegraphics[width=0.23\textwidth]{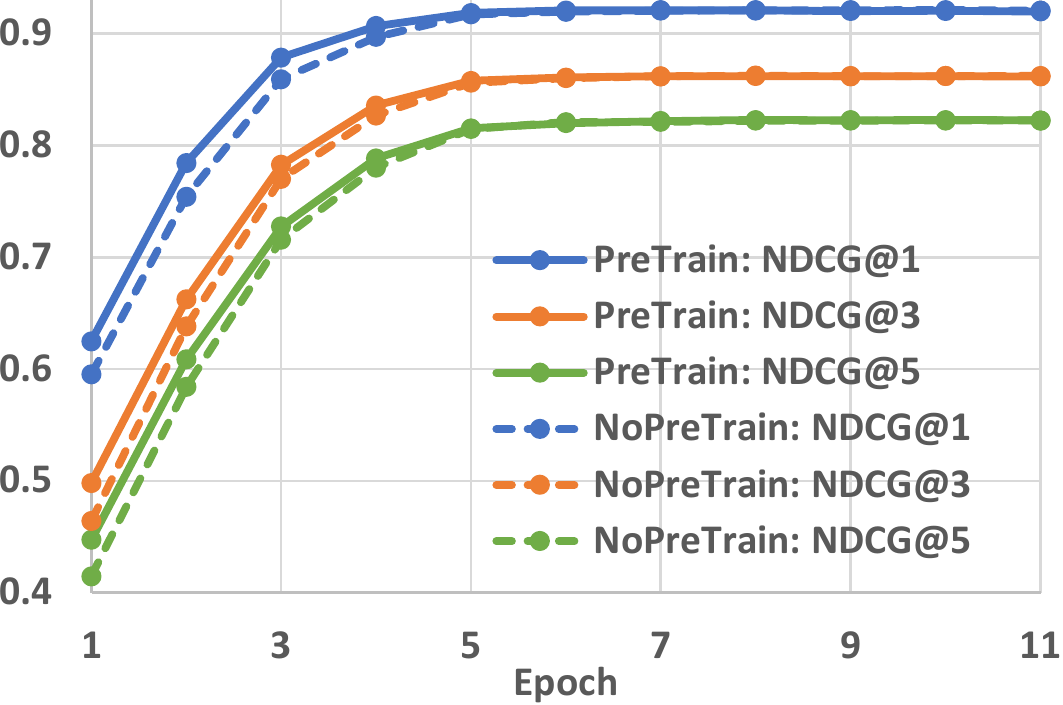}}
\subfigure[PubMed, Training Loss]{
\includegraphics[width=0.23\textwidth]{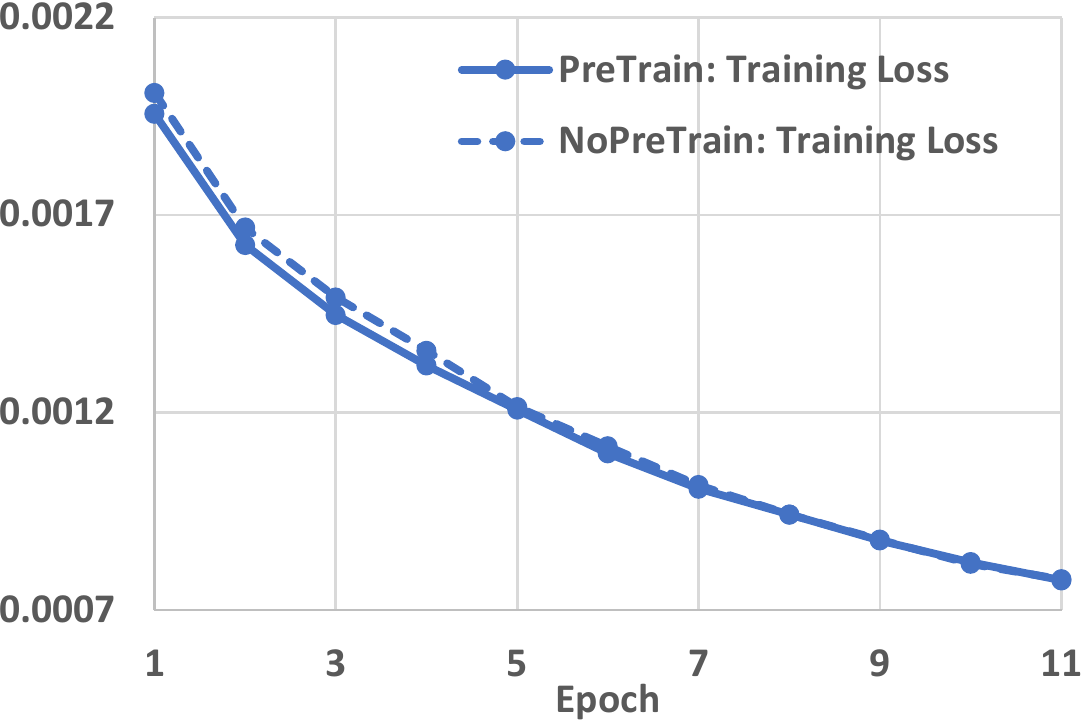}}
\hspace{-0.25em}
\subfigure[PubMed, Validation NDCG@$k$]{
\includegraphics[width=0.23\textwidth]{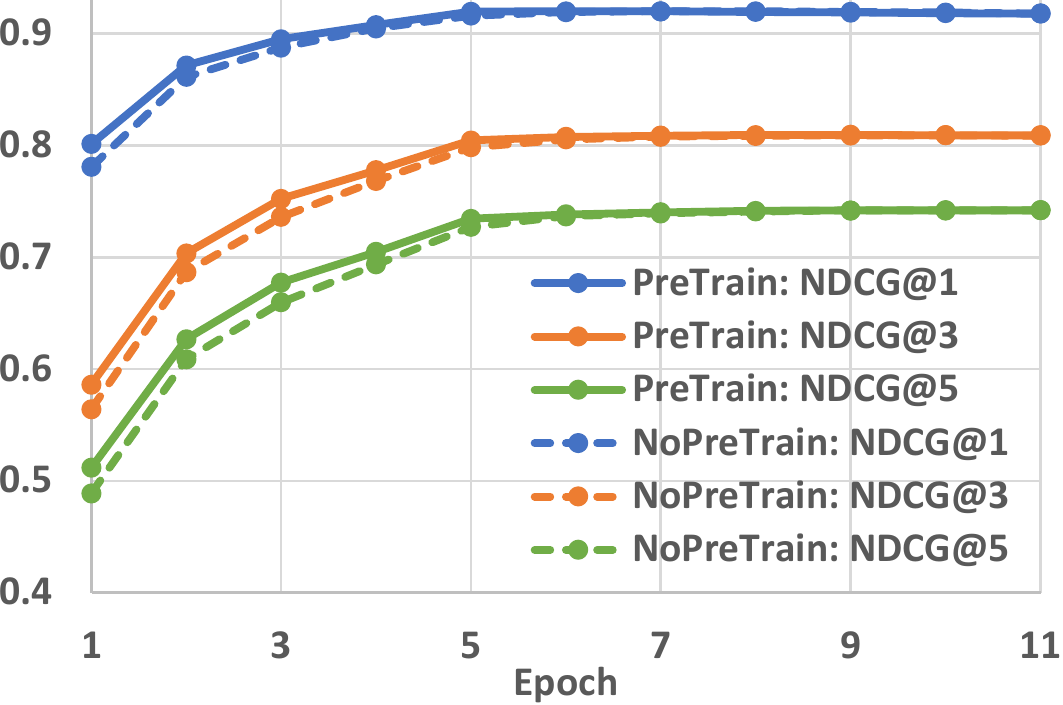}}
\caption{Performance of \textsf{\model} during the training process with and without metadata-aware embedding pre-training.} 
\vspace{-1em}
\label{fig:pretrain}
\end{figure}
We have shown the positive contribution of leveraging different types of metadata in \textsf{\model}, which is a combined effect of metadata-aware embedding pre-training and metadata-aware Transformer encoding. Now we would like to show the advantages of embedding pre-training alone.
To facilitate this, we create another ablation version, \textbf{\textsf{\model}-NoPreTrain}, which bypasses metadata-aware embedding pre-training and directly uses GloVe.6B.100d as initialized embeddings of our neural classifier. 

Figure \ref{fig:pretrain} demonstrates the performance of \textsf{\model} and \textsf{\model}-NoPreTrain during the training process. 
The x-axis represents training epochs. In Figures \ref{fig:pretrain}(a) and (c), the y-axis is the average training loss of the last 100 batches in epoch $x$. In Figures \ref{fig:pretrain}(b) and (d), the y-axis represents NDCG@$k$ ($k=1,3,5$) of the trained classifier on the validation set after epoch $x$.
The full model is denoted by solid lines and NoPreTrain is denoted by dashed lines. 
We can observe that: 
(1) In earlier epochs, the full model achieves evidently higher NDCG@$k$ scores and lower training loss than NoPreTrain, indicating that embedding pre-training provides a warm start to neural classifier training. This is intuitive because the embeddings of metadata instances and words unseen in GloVe.6B.100d need to be randomly initialized in \textsf{\model}-NoPreTrain. As training proceeds, the performance of NoPreTrain becomes on par with that of the full model, which means metadata-aware pre-training cannot significantly boost the final NDCG@$k$ scores. The reason could be that our Transformer-based encoder already captures higher-order information than the pre-training step does (as mentioned in Section \ref{sec:trans}), which makes up for the cold start caused by random initialization.
(2) The NDCG@$k$ curves of both models converge before epoch 11. On MAG-CS, the full model achieves its best NDCG@1 at epoch 7 while NoPreTrain gets the highest NDCG@1 at epoch 10. On PubMed, the peak NDCG@1 scores of \textsf{\model} and \textsf{\model}-NoPreTrain are at epoch 7 and epoch 8, respectively. 
To summarize, on both datasets, the full model converges earlier than NoPreTrain in terms of precision on the validation set. In other words, metadata-aware pre-training increases the speed of model convergence in \textsf{\model}.

\subsection{Case Study}

\begin{table}[]
\caption{Case Study on MAG-CS. {\color{orange} Orange}: Incorrect predictions. {\color{blue} Blue}: Correct predictions when utilizing metadata, and the corresponding signals. {\color{darkg} Green}: Correct predictions when utilizing the hierarchy, and the corresponding signals.}
\vspace{-0.5em}
\footnotesize
\begin{tabular}{p{8.1cm}}
\hline
\cellcolor{black!20} \textbf{\small Case 1: Effect of Metadata} \\
\hline
\textbf{Title}: Improving Text Categorization Methods for Event Tracking                                                                                       \\
\textbf{Venue}: {\color{blue} \textbf{SIGIR}} (2000)                                \\
\textbf{Authors}: {\color{blue} \textbf{Yiming Yang}}, Tom Ault, Thomas Pierce, Charles W. Lattimer                                                                                   \\ \hline
\textbf{Abstract}: Automated tracking of events from chronologically ordered document streams is a new challenge for statistical text classification. Existing learning techniques must be adapted or improved in order to effectively handle difficult situations where the number of positive training instances per event is extremely small, the majority of training documents are unlabelled, and most of the events have a short duration in time. We adapted several supervised text categorization methods, specifically several new variants of the k-Nearest Neighbor (kNN) algorithm ...                                                          \\ \hline \hline
\textbf{Ground Truth Labels}: \texttt{Data Mining, Machine Learning, Information Retrieval, K Nearest Neighbors Algorithm, Pattern Recognition
} \\ \hline
\textbf{Top-5 Predictions of Transformer}: \texttt{K Nearest Neighbors Algorithm (\cmark), Data Mining (\cmark), Pattern Recognition (\cmark), Machine Learning (\cmark), {\color{orange} Nearest Neighbor Search (\xmark)}}     \\ \hline
\textbf{Top-5 Predictions of \textsf{\model}-NoHierarchy}:   \texttt{K Nearest Neighbors Algorithm (\cmark), Data Mining (\cmark), Pattern Recognition (\cmark), {\color{blue} Information Retrieval (\cmark)}, Machine Learning (\cmark)}                                                                                                                        \\ \hline
\textbf{Top-5 Predictions of \textsf{\model}}: \texttt{K Nearest Neighbors Algorithm (\cmark), Data Mining (\cmark), {\color{blue} Information Retrieval (\cmark)}, Pattern Recognition (\cmark), Machine Learning (\cmark)}                                                     \\ \hline
\hline
\cellcolor{black!20} \textbf{\small Case 2: Effect of Hierarchy} \\
\hline
\textbf{Title}: Automatic Derivation of a Phoneme Set with Tone Information for Chinese Speech Recognition Based on Mutual Information Criterion                                                                                     \\
\textbf{Venue}: ICASSP (2006)                                \\ \hline
\textbf{Abstract}: An appropriate approach to model tone information is helpful for building Chinese large vocabulary continuous speech recognition system. We propose to derive an efficient phoneme set of tone-dependent sub-word units to build a recognition system, by iteratively merging a pair of tone-dependent units according to the principle of minimal loss of the mutual information. The mutual information is measured between the word tokens and their phoneme transcriptions in a training text corpus, based on the system lexical and language model. ...                                              \\ \hline
\textbf{Hypernymy Information}: \textbf{{\color{darkg} parents( \texttt{Language Model} )}} $=$ \{ \texttt{Artificial Intelligence, Speech Recognition, \textbf{{\color{darkg} Natural Language Processing}}} \}
\\ \hline \hline
\textbf{Ground Truth Labels}: \texttt{Vocabulary, Homophone, Natural Language, Audio Mining, Speech Recognition, Natural Language Processing, Word Error Rate, Language Model, Text Corpus, Pattern Recognition, Mutual Information} \\ \hline
\textbf{Top-5 Predictions of Transformer}: \texttt{Speech Recognition (\cmark), {\color{orange} Discriminative Model (\xmark)}, Language Model (\cmark), Mutual Information (\cmark), Vocabulary (\cmark)}     \\ \hline
\textbf{Top-5 Predictions of \textsf{\model}-NoHierarchy}:   \texttt{Mutual Information (\cmark), Speech Recognition (\cmark), Vocabulary (\cmark), {\color{orange} Discriminative Model (\xmark)}, Language Model (\cmark)}                                                                                                                        \\ \hline
\textbf{Top-5 Predictions of \textsf{\model}}: \texttt{Text Corpus (\cmark), Speech Recognition (\cmark), Language Model (\cmark), Mutual Information (\cmark), {\color{darkg} Natural Language Processing (\cmark)}}                                                     \\ \hline
\hline
\cellcolor{black!20} \textbf{\small Case 3: An Error of \textsf{\model}} \\
\hline
\textbf{Title}: The Winograd Schema Challenge and Reasoning about Correlation                                                                                \\
\textbf{Venue}: AAAI (2015)                                \\
\hline
\textbf{Abstract}: The Winograd Schema Challenge is an alternative to the Turing Test that may provide a more meaningful measure of machine intelligence. It poses a set of coreference resolution problems that cannot be solved without human-like reasoning. In this paper, we take the view that the solution to such problems lies in establishing discourse coherence. Specifically, we examine two types of rhetorical relations that can be used to establish discourse coherence: positive and negative correlation. We introduce a framework for reasoning about correlation ...                                                   \\ \hline \hline
\textbf{Ground Truth Labels}: \texttt{Coreference, Artificial Intelligence, Natural Language Processing, Winograd Schema Challenge, Turing Test
} \\ \hline
\textbf{Top-5 Predictions of Transformer}: \texttt{Turing Test (\cmark), Winograd Schema Challenge (\cmark), Natural Language Processing (\cmark), Coreference (\cmark), Artificial Intelligence (\cmark)} \\ \hline
\textbf{Top-5 Predictions of \textsf{\model}-NoHierarchy}: \texttt{Winograd Schema Challenge (\cmark), Turing Test (\cmark), Coreference (\cmark), {\color{orange} Machine Learning (\xmark)}, Artificial Intelligence (\cmark)}                                                                                                                        \\ \hline
\textbf{Top-5 Predictions of \textsf{\model}}: \texttt{Turing Test (\cmark), Winograd Schema Challenge (\cmark), Coreference (\cmark), {\color{orange} Machine Learning (\xmark)}, Artificial Intelligence (\cmark)}                                          \\ \hline
\end{tabular}
\label{tab:case}
\end{table}

We now conduct case studies to qualitatively understand the effects of incorporating metadata and the label hierarchy. Table \ref{tab:case} compares the full \textsf{\model} model with \textsf{\model}-NoHierarchy and Transformer on the predictions of three MAG-CS papers. For each paper, we show its text, (part of) metadata/hierarchy information, ground truth labels as well as top-5 predicted labels of the three compared approaches. Recall that \textsf{\model}-NoHierarchy does not use any label hierarchy information, and Transformer is unaware of both metadata and the hierarchy.

In Case 1, the paper has a ground truth label ``\texttt{Information Retrieval}''. Although the term ``retrieval'' does not explicitly appear in the title and abstract, metadata signals (especially the venue ``\textit{SIGIR}'' and one of the authors ``\textit{Yiming Yang}'') successfully indicate the paper's relevance to ``\texttt{Information Retrieval}''. However, Transformer fails to predict ``\texttt{Information Retrieval}'' in its top-5 choices as it is blind to metadata. Instead, it makes a wrong prediction ``\texttt{Nearest Neighbor Search}''. In contrast, both \textsf{\model} and \textsf{\model}-NoHierarchy can observe metadata information, thus both of them correctly pick ``\texttt{Information Retrieval}''.

In Case 2, the paper is related to a fine-grained topic ``\texttt{Language Model}'' and a broader category ``\texttt{Natural Language Processing}''. As the paper mentions ``language model'' and related terms in its abstract, the three compared approaches all include ``\texttt{Language Model}'' correctly in their top-5 choices. According to the hypernymy information, we can see three parent categories of ``\texttt{Language Model}'', which are ``\texttt{Artificial Intelligence}'', ``\texttt{Speech Recognition}'', and ``\texttt{Natural Language Processing}''. The last two are in the ground truth labels of this paper. Unlike ``\texttt{Speech Recognition}'' which can be easily inferred from the title, ``\texttt{Natural Language Processing}'' can be neither found in the text nor indicated by the venue. Therefore, ``\texttt{Natural Language Processing}'' is missed by Transformer and \textsf{\model}-NoHierarchy. In contrast, by observing the hierarchy information, \textsf{\model} successfully picks ``\texttt{Natural Language Processing}'' in its top-5 predictions.

Case 1 and Case 2 reflect the benefit of considering metadata and the hierarchy, respectively. However, in a few cases, such additional signals may also confuse our model. We show an error made by \textsf{\model} in Case 3. The paper is about the Winograd Schema Challenge. Transformer successfully predicts all ground truth labels in its top-5 choices. However, both \textsf{\model}-NoHierarchy and \textsf{\model} give a wrong prediction ``\texttt{Machine Learning}'', probably because the paper is published at AAAI which has many machine learning studies. In fact, the paper is purely based on formal logical reasoning and has no machine learning related component. This case implies an interesting future direction on how to automatically select topic-indicative metadata instances to help classification.

\section{Related Work}
\noindent \textbf{Multi-label Text Classification.} 
Traditional multi-label text classification approaches mainly use bag-of-words representations and can be divided into three categories: (1) \textit{One-vs-all} methods \cite{babbar2017dismec,yen2016pd,yen2017ppdsparse} exploit data sparsity to learn a classifier for each label independently. (2) \textit{Tree-based} approaches \cite{jain2016extreme,prabhu2014fastxml,prabhu2018extreme,siblini2018craftml} recursively partition the feature space at each non-leaf node and learn a classifier focusing on only a few active labels at each leaf node. (Note that they are hierarchically partitioning the feature space instead of the label space, thus cannot be viewed as conventional hierarchical text classification methods.) (3) \textit{Embedding-based} approaches \cite{bhatia2015sparse,guo2019breaking,chen2012feature,tagami2017annexml} represent labels as low-dimensional vectors and perform classification by finding the nearest label neighbors of each document in the latent space.
Recently, deep learning based methods leverage deep neural architectures to learn better text representations. For example, Liu et al. \cite{liu2017deep} propose a convolutional neural network with dynamic pooling and a hidden bottleneck layer for text encoding. Nam et al. \cite{nam2017maximizing} leverage recurrent neural networks to encode text sequences and generate predicted labels sequentially. You et al. \cite{you2019attentionxml} adopt attention models to capture the most relevant parts of the input text to each label. Chang et al. \cite{chang2020taming} utilize pre-trained Transformers as neural matchers to perform classification. There are also multi-label classifiers specifically designed for biomedical literature such as DeepMeSH \cite{peng2016deepmesh}, MeSHProbeNet \cite{xun2019meshprobenet}, and FullMeSH \cite{dai2020fullmesh}, where the task is named as MeSH indexing. However, all these models are designed for a flat label space and do not consider the hierarchical dependencies and intercorrelation between labels, while our \textsf{\model} introduces hypernymy guided regularization.

\vspace{1mm}

\noindent \textbf{Hierarchical Text Classification.} 
Hierarchical text classification aims to leverage label hierarchies to improve classification performance. Early approaches such as Hierarchical SVM \cite{dumais2000hierarchical,liu2005support} assume the hierarchy has a tree structure and adopt a top-down training strategy. 
In contrast, bottom-up methods \cite{bennett2009refined} backpropagate the labels from the leaves to the top layer. To further exploit the parent-child relationships between labels, Gopal and Yang \cite{gopal2013recursive,gopal2015hierarchical} introduce a recursive regularization to encourage the similarity between child classifiers and their parent classifier. Peng et al. \cite{peng2018large} further extend this regularization to graph neural networks. Wehrmann et al. \cite{wehrmann2018hierarchical} combine the ideas of training a local classifier per level and adopting global optimization techniques to mitigate exposure bias. Huang et al. \cite{huang2019hierarchical} further improve Wehrmann et al.'s model by introducing label attention per level. The global structure of hierarchies is also used in various models by other studies, such as meta-learning \cite{wu2019learning}, reinforcement learning \cite{mao2019hierarchical} and tree/graph based neural networks \cite{zhou2020hierarchy}. However, all approaches mentioned above only consider classifying plain text sequences. For documents with rich metadata information, our \textsf{\model} uses pre-training and attention mechanisms to make full use of metadata.

\vspace{1mm}

\noindent \textbf{Metadata-Aware Text Classification.} Some previous studies try to incorporate metadata information for specific classification tasks. For example, Tang et al. \cite{tang2015learning} leverage user and product information for review sentiment analysis. Zhang et al. \cite{zhang2017rate} employ user biography data for tweet localization. Zhang et al. \cite{zhang2019higitclass} use both the creator and the repository tags for GitHub repository classification. To solve general classification tasks, Kim et al. \cite{kim2019categorical} inject categorical metadata signals into a deep neural classifier as additional features. There are also studies considering weakly supervised settings. Zhang et al. \cite{zhang2020minimally,zhang2021hierarchical} propose to generate synthesized training samples with the help of metadata-aware representation learning. Mekala et al. \cite{mekala2020meta} incorporate metadata as additional supervision for text classification with seed words only. However, in these studies, each document is assigned to only one category, and the label space is usually small.

\section{Conclusion and Future Work}
We present \textsf{\model}, a multi-label text classification framework that simultaneously leverages metadata and label hierarchy signals. The framework is featured by a metadata-aware embedding pre-training module, a metadata-aware Transformer encoder, and a hypernymy regularization module. The pre-training module learns better text and metadata representations by characterizing their relationships in a joint embedding space. The Transformer encoder facilitates higher-order interactions between words and metadata. The hypernymy regularization terms model the similarity and the inclusive relationship between parent and child categories. Experimental results demonstrate the superiority of \textsf{\model} towards competitive baselines. Moreover, we validate the contribution of incorporating metadata and label hierarchy through ablation analysis and case studies.

There are several future directions in light of our model design and experiments. First, it is interesting to study the contribution of various metadata in different domains (e.g., product reviews, encyclopedia webpages, etc.) and how to automatically select the metadata that is helpful to the classification task. Second, we may look for more complicated document encoder architectures that can consider the types of metadata as well as the hierarchy information.

\begin{acks}
Research was sponsored in part by US DARPA KAIROS Program No. FA8750-19-2-1004 and SocialSim Program No. W911NF-17-C-0099, National Science Foundation IIS-19-56151, IIS-17-41317, IIS 17-04532, and IIS 16-18481, and DTRA HDTRA11810026. Any opinions, findings, and conclusions or recommendations expressed herein are those of the authors and should not be interpreted as necessarily representing the views, either expressed or implied, of DARPA or the U.S. Government.
We thank Xiaodong Liu and Boya Xie for insightful discussions on this project and anonymous reviewers for valuable feedback.
\end{acks}

\balance
\bibliography{www}

\end{document}